\documentclass[conference]{IEEEtran}
\IEEEoverridecommandlockouts
\usepackage{cite}
\usepackage{amsmath,amssymb,amsfonts}
\usepackage{algorithmic}
\usepackage{graphicx}
\usepackage{textcomp}
\usepackage{xcolor}
\usepackage{url}
\usepackage[hidelinks]{hyperref}

\def\BibTeX{{\rm B\kern-.05em{\sc i\kern-.025em b}\kern-.08em
    T\kern-.1667em\lower.7ex\hbox{E}\kern-.125emX}}

\ifCLASSOPTIONcompsoc
\usepackage[caption=false,font=normalsize,labelfon
t=sf,textfont=sf]{subfig}
\else
\usepackage[caption=false,font=footnotesize]{subfi
	g}
\fi

\hypersetup{
	colorlinks=false
}

\graphicspath{{./figs/}}

\begin{document}
	
\urlstyle{tt}

\title{Sim-to-real reinforcement learning applied to end-to-end vehicle control \\
\thanks{\IEEEauthorrefmark{2} and \IEEEauthorrefmark{3} contributed equally to this work}
} 

\author{\IEEEauthorblockN{András Kalapos\IEEEauthorrefmark{1}, Csaba Gór\IEEEauthorrefmark{2}, Róbert Moni\IEEEauthorrefmark{3} and István Harmati\IEEEauthorrefmark{4}}
\IEEEauthorblockA{\IEEEauthorrefmark{1}\textit{BME, Dept. of Control Engineering and Information Technology}, Budapest, Hungary, andras.kalapos.research@gmail.com}
\IEEEauthorblockA{\IEEEauthorrefmark{2}\textit{Continental ADAS AI}, Budapest, Hungary, csaba.gor@continental.com}
\IEEEauthorblockA{\IEEEauthorrefmark{3}\textit{BME, Dept. of Telecommunications and Media Informatics}, Budapest, Hungary,  robertmoni@tmit.bme.hu}
\IEEEauthorblockA{\IEEEauthorrefmark{4}\textit{BME, Dept. of Control Engineering and Information Technology}, Budapest, Hungary, harmati@iit.bme.hu}
}

\maketitle

\begin{abstract}
In this work, we study vision-based end-to-end reinforcement learning on vehicle control problems, such as lane following and collision avoidance. Our controller policy is able to control a small-scale robot to follow the right-hand lane of a real two-lane road, while its training was solely carried out in a simulation. Our model, realized by a simple, convolutional network, only relies on images of a forward-facing monocular camera and generates continuous actions that directly control the vehicle. To train this policy we used Proximal Policy Optimization, and to achieve the generalization capability required for real performance we used domain randomization. We carried out thorough analysis of the trained policy, by measuring  multiple performance metrics and comparing these to baselines that rely on other methods. To assess the quality of the simulation-to-reality transfer learning process and the performance of the controller in the real world, we measured simple metrics on a real track and compared these with results from a matching simulation. Further analysis was carried out by visualizing salient object maps.
	
\end{abstract}


\section{Introduction}
Reinforcement learning has been used to solve many control and robotics tasks, however, only a handful of papers has been published so far that apply this technique to end-to-end driving \cite{AsyncMethods,EndToEndRaceDriving,LearningToDriveInADay,SSINet,DeepRacer,Szemenyei2019}. Even fewer works focus on reinforcement learning-based driving, trained only in simulations but applied to real-world problems. Generally, bridging the gap between simulation and the real world is an important transfer learning problem related to reinforcement learning and is an unresolved task for researchers. 

%Mnih et al. \cite{AsyncMethods} primarily proposes new RL algorithms but also analyses the performance of these on the TORCS simulator by training policies to predict discrete control actions based on a single image of a forward-facing camera. 

Mnih~et~al.~\cite{AsyncMethods} proposed a method to train vehicle controller policies that predict discrete control actions based on a single image of a forward-facing camera. 
Jaritz~et~al.~\cite{EndToEndRaceDriving} used WRC6, a realistic racing simulator to train a vision-based road following policy. They assessed the policy's generalization capability by testing on previously unseen tracks and on real diving videos, in an open-loop configuration, but their work didn’t extend to evaluation on real vehicles in closed-loop control. 
Kendall~et~al.~\cite{LearningToDriveInADay} demonstrated real-world driving by training a lane-following policy exclusively on a real vehicle, under the supervision of a safety driver. 
Shi~et~al.~\cite{SSINet} presented research that involves training reinforcement learning agents in Duckietown similarly to ours, however, they mainly focused on presenting a method that explains the reasoning of the trained agents, rather than on the training methods. 
Similarly to our research, Balaji~et~al.~\cite{DeepRacer} presented a method for training a road-following policy in a simulator using reinforcement learning and tested the trained agent in the real world, yet their primary contribution is the DeepRacer platform, rather than the in-depth analysis of the road following policy.

In this contribution, we study vision-based end-to-end reinforcement learning on vehicle control problems and propose a solution that performs lane following in the real world, using continuous actions, without any real data provided by an expert (as~in~\cite{LearningToDriveInADay}). Also, we perform validation of the trained policies in both the real and simulated domains.

%In this work, we study vision-based end-to-end reinforcement learning on vehicle control problems, such as lane-following and collision avoidance. Moreover, the simulation to reality transfer of these control problems is also a central question of our research.

\section{Methods}
\begin{figure}[tbp]
	\centering
	\subfloat[Simulated]{\includegraphics[width=2.5cm]{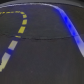}
		\label{fig_sal_obj_sim}}
	\subfloat[Real]{\includegraphics[width=2.5cm]{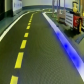}
		\label{fig_sal_obj_real}}
	\subfloat[Collision avoidance]{\includegraphics[width=2.5cm]{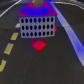}
		\label{fig_sal_obj_lfv}}
	\caption{Salient objects highlighted on observations in different domains and tasks. Blue regions represent high activations throughout the network.}
	\label{fig_sal_obj}
\end{figure}
We trained a neural network-based controller that takes images from a forward-looking monocular camera and produces control signals to drive a vehicle in the right lane of a two-way road. The vehicle to be controlled is a small differential-wheeled mobile robot, a so-called Duckiebot, which is part of the Duckietown ecosystem\cite{Duckietown}, a simple and accessible platform for research and education on mobile robotics and autonomous vehicles. The primary objective is to travel as far as possible under a given time, without leaving the road (while lane departure is allowed, but not preferred). 

Training and evaluation code for this paper will be open sourced after the 5\textsuperscript{th} AI-Driving Olympics and will be available on GitHub\footnote{\url{https://github.com/kaland313/Duckietown-RL}}.

\subsection{Reinforcement learning algorithm}
To train the policy we used Proximal Policy Optimization algorithm \cite{PPO} for its stability, sample-complexity, and ability to take advantage of multiple parallel workers.

Policy optimization algorithms are on-policy reinforcement learning methods that directly update the $\pi_\theta(a_t|s_t)$ policy based on the $a_t$ actions and the $r_t$ reward received for them ($\theta$ denotes the trainable parameters of the policy and $s_t$ is the observation at timestep $t$). The policy used for these algorithms is stochastic and in case of deep reinforcement learning it's implemented by a neural network, which is updated using gradient methods. In simpler versions of the algorithm (such as REINFORCE\cite{REINFORCE}), the gradients are estimated by $\hat{g} = \mathop{\mathbb{\hat{E}}}_{\tau\sim \pi_\theta}\left [ \nabla_\theta \log \pi_\theta(a_t|s_t) {G}^{\pi_\theta}(a_t,s_t)\right ]$, where $G_t$ is the return.

Proximal Policy Optimization performs the weight updates using a special loss function to keep the new policy close to the old, thereby improving the stability of the training. Two loss functions were proposed by Schulman et al.~\cite{PPO}: 

\small\begin{equation}
	\mathfrak{L}_{CLIP}(\theta)=\mathop{\mathbb{\hat{E}}}\left [\min \left( r_t(\theta)  \hat{A}_t, \mathrm{clip} (r_t(\theta), 1-\epsilon, 1+\epsilon)\hat{A}_t\right) \right]
\end{equation}
\begin{equation}
	\mathfrak{L}_{KLPEN}(\theta)=\mathop{\mathbb{\hat{E}}}\left[r_t(\theta) \hat{A} - \beta \mathrm{KL}\left[\pi_{\theta_{old}}(\cdot|s_t), \pi_{\theta}(\cdot|s_t) \right] \right] 
\end{equation}
\normalsize
where $\mathrm{clip}(\cdot)$ and $\mathrm{KL}[\cdot]$ refer to the clipping function and KL-divergence respectively, while $\hat{A}$ is calculated as the generalized advantage estimate~\cite{GAE}. In these loss functions $r_t(\theta) = \frac{\pi_\theta(a_t|s_t)}{\pi_{\theta_{old}}(a_t|s_t)}$, $\epsilon$ is a constant usually in the $[0.1, 0.3]$ range, while $\beta$ is an adaptive parameter. We used an open-source implementation of the algorithm\cite{RLlib}, which performs the gradient updates based on the weighted sum of these loss functions.
 
\subsection{Policy architecture}
\begin{figure}[tbp]
	\centering
	\includegraphics[width=8.5cm]{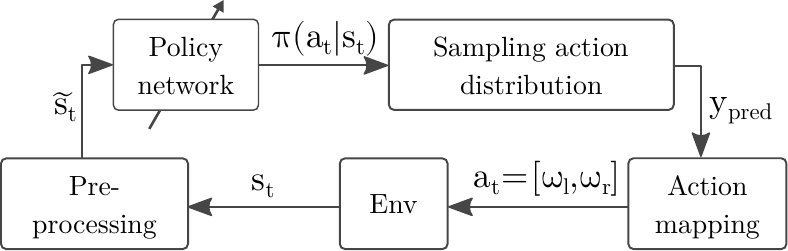}
	\caption{Illustration of the policy architecture with the used notations.}
	\label{fig:policyarchitecture}
\end{figure}

The controller policy is realized by a shallow (4-layer) convolutional neural network. We consider this policy end-to-end because the only learning component is the neural network, which directly computes actions based on observations from the environment. Both the policy and the value network use the architecture presented by Mnih et al. \cite{AsyncMethods}, with no weight sharing (with the only difference of using linear activation on the output of the policy network). 

Some pre- and post-processing is applied to the observations and actions respectively, but these only perform very simple transformations. The input of the policy network is the last three observations (images) scaled, cropped and stacked (along the depth axis). The observations returned by the environment are  $640\times480$ (width, height) RGB images whose top third mainly shows the sky, therefore is cropped. Then, the cropped images are scaled down to $84\times84$ resolution (note the uneven scaling), which are then stacked along the depth axis resulting in $84\times84\times9$ input tensors. The last three images are stacked to provide the policy with information about the robot's speed and acceleration. 

We experimented with multiple action representations (see sec.~\ref{sec_action_representations}), depending on these the policy outputs one or two scalar values which control the vehicle. The policy is stochastic, therefore the output of the neural network produces the parameters of a (multivariate diagonal) normal distribution, which is sampled to acquire actions.

\subsection{Action representations} \label{sec_action_representations}
The vehicle to be controlled is a differential-wheeled robot, therefore the most general action representation is to directly predict the angular velocities of the two wheels as continuous values in the $\omega_{l,r}\in[-1;1]$ range (where 1 and -1 correspond to rotating forward and backward at full speed). However, this action space allows for actions that are not necessary for the maneuvers we examine in this paper. Moreover, by allowing unnecessary actions, the reinforcement learning algorithm must rule these out, potentially making the exploration of the action space more difficult therefore increasing the steps required to train an agent. Several methods can be used to constrain and simplify the action space, such as discretization, clipping some actions, or mapping to a lower-dimensional space.

Most previous works (\hspace{-0.01mm}\cite{AsyncMethods,EndToEndRaceDriving,DeepRacer}) use discrete action spaces, thus the neural network in these policies selects one from a set of hand-crafted actions (steering, throttle combinations), while Kendall et al. \cite{LearningToDriveInADay} utilize continuous actions, as we do. However, they don't predict throttle directly, only a speed set-point for a classical controller.
 
In order to test the reinforcement learning algorithm's ability to solve the most general problem, we experimented with multiple action mappings and simplifications of the action space. These were
\subsubsection{Wheel Velocity} The policy directly outputs wheel velocities $\omega_{l,r}\in[-1;1]$ 
\subsubsection{Wheel Velocity - Positive Only} Only allow positive wheel velocities, because only these are required to move forward. Values predicted outside the $\omega_{l,r}\in[0;1]$ interval are clipped.
\subsubsection{Wheel Velocity - Braking} Wheel velocities could still only fall in the $\omega_{l,r}\in[0;1]$ interval, but the predicted values are interpreted as the amount of braking from the maximum speed $\omega_{l,r}=1-y_{pred,l,r}$. The main differentiating factor from the Positive Only option is the bias towards moving forward at full speed. 
\subsubsection{Steering} Predicting a  scalar value that is continuously mapped to combinations of wheel velocities. The 0.0 scalar value corresponds to going straight (at full speed), while -1.0 and 1.0 refer to turning left or right, with one wheel completely stopped and the other one going at full speed. The speed of the robot is always maximal for a particular steering value.  

\subsection{Reward shaping}
\begin{figure}[tbp]
	\centering
	\includegraphics[width=8cm]{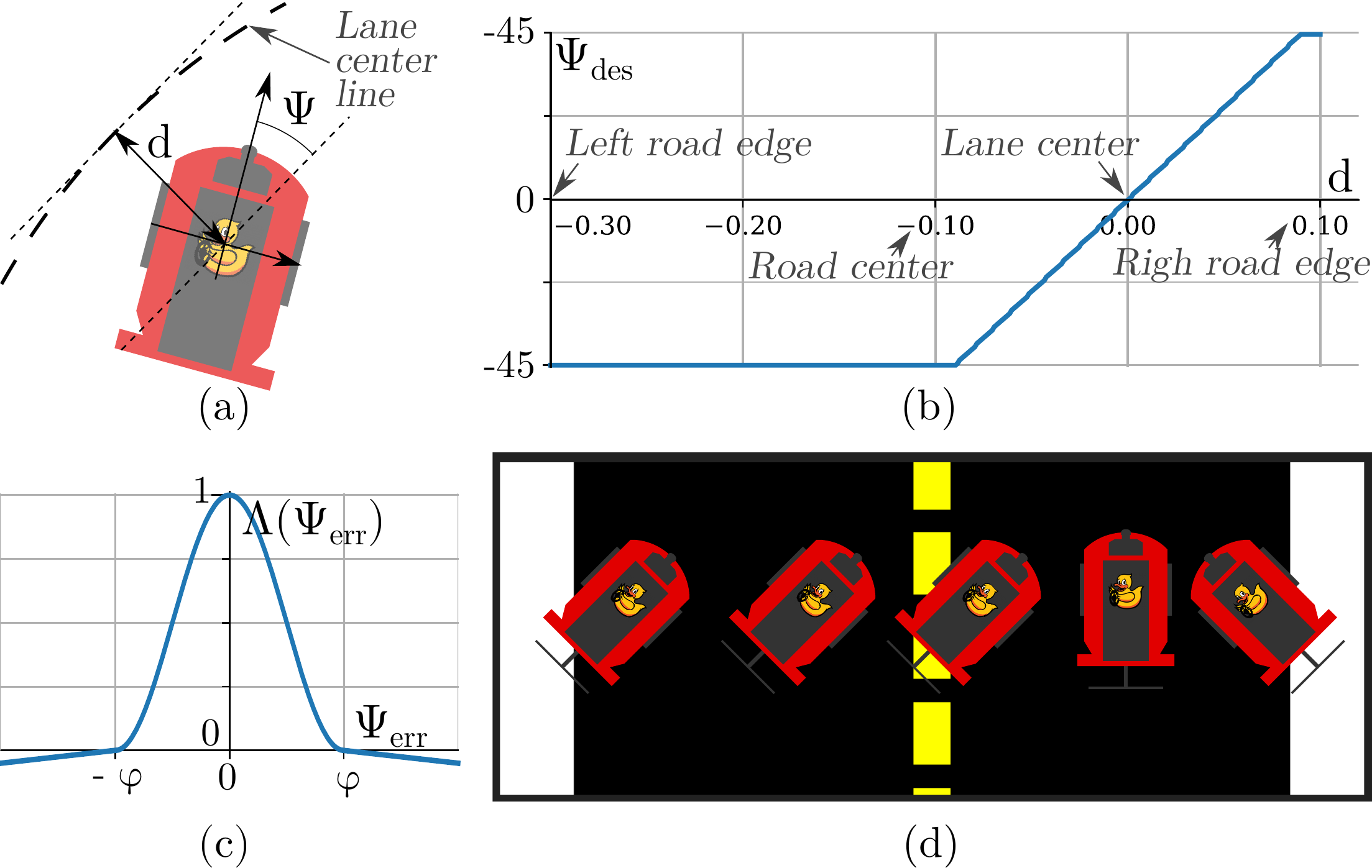}
	\caption[]{Explanation of the proposed Orientation reward. (a) explains $\Psi, d$, (b) shows how the desired orientation depends on the lateral error, (d), shows some examples of desired configurations, while (c) shows the function \\
		\begin{minipage}{\columnwidth} 
		\[ 
			\Lambda (x) = \left\{\begin{array}{ll}
			\frac{1}{2} + \frac{1}{2}\cos \left(\pi \frac{x}{\varphi}\right) & \textrm{if } -1\ge x\ge 1\\ 
			\varepsilon (-|\frac{x}{\varphi}| + 1)  & \textrm{otherwise }
			\end{array}\right., \varepsilon \in [10^{-1},10^{-2}] \tag{3} \label{eq_Lambda}
		\]
		\end{minipage}
 	}
	\label{fig_reward}
\end{figure}
The reward function is a fundamental element of every reinforcement learning problem as it serves the important role of converting a task from a textual description to a mathematical optimization problem. The primary objective for the agent is to travel as far as possible under a given time in the right lane, therefore we propose two rewards that promote this behavior. 
\subsubsection{Distance traveled} The agent is directly rewarded proportionally to the distance it moved further along the right lane under every step. Only longitudinal motion is counted, and only if the robot stayed in the right lane.

\subsubsection{Orientation} The agent is rewarded if it is facing towards and moves in a certain desired orientation, which is determined based on its lateral position. In simple terms, it is rewarded the most if it faces towards the center of the right lane (some example configurations are shown on fig.~\ref{fig_reward}d). A term proportional to the angular velocity of the faster moving wheel is also added to encourage fast motion. 

This reward is calculated as $r = \lambda_{\Psi} r_{\Psi}(\Psi, d) + \lambda_v r_v(\omega_l,\omega_r)$, where $r_{\Psi}(\cdot), r_v(\cdot)$ are the orientation and velocity based components, while $\lambda_{\Psi}, \lambda_v$ constants scale these to $[-1,1]$. $\Psi, d$ are orientation and lateral error from the desired trajectory, which is the center line of the right lane (see fig.~\ref{fig_reward}a).% $\omega_l,\omega_r$ are wheel angular velocities.

The orientation-based term is calculated as $r_{\Psi}(\Psi, d) = \Lambda(\Psi_{err}) = \Lambda(\Psi-\Psi_{des}(d))$, where $\Psi_{des}(d)$ is the desired orientation, calculated based on the lateral distance from the desired trajectory (see fig.~\ref{fig_reward}b for the illustration of $\Psi_{des}(d)$). The $\Lambda$ function achieves that $|\Psi_{err}| < \varphi$ error is promoted largely, while error larger than this leads to small negative reward (formal description \eqref{eq_Lambda} and a plot of $\Lambda$ is shown on fig.~\ref{fig_reward}c). $\varphi=50^\circ$ hyper-parameter was selected arbitrarily.

The velocity-based component is calculated as $r_v(\omega_l,\omega_r) = \mathrm{max}(\omega_l,\omega_r)$ to reward high speed motion equally in straight and curved sections, where only the outer wheel can rotate as fast as on straight sections.

\subsection{Simulation to reality transfer}

\begin{figure}[tbp]
	\centering
	\includegraphics[width=8cm]{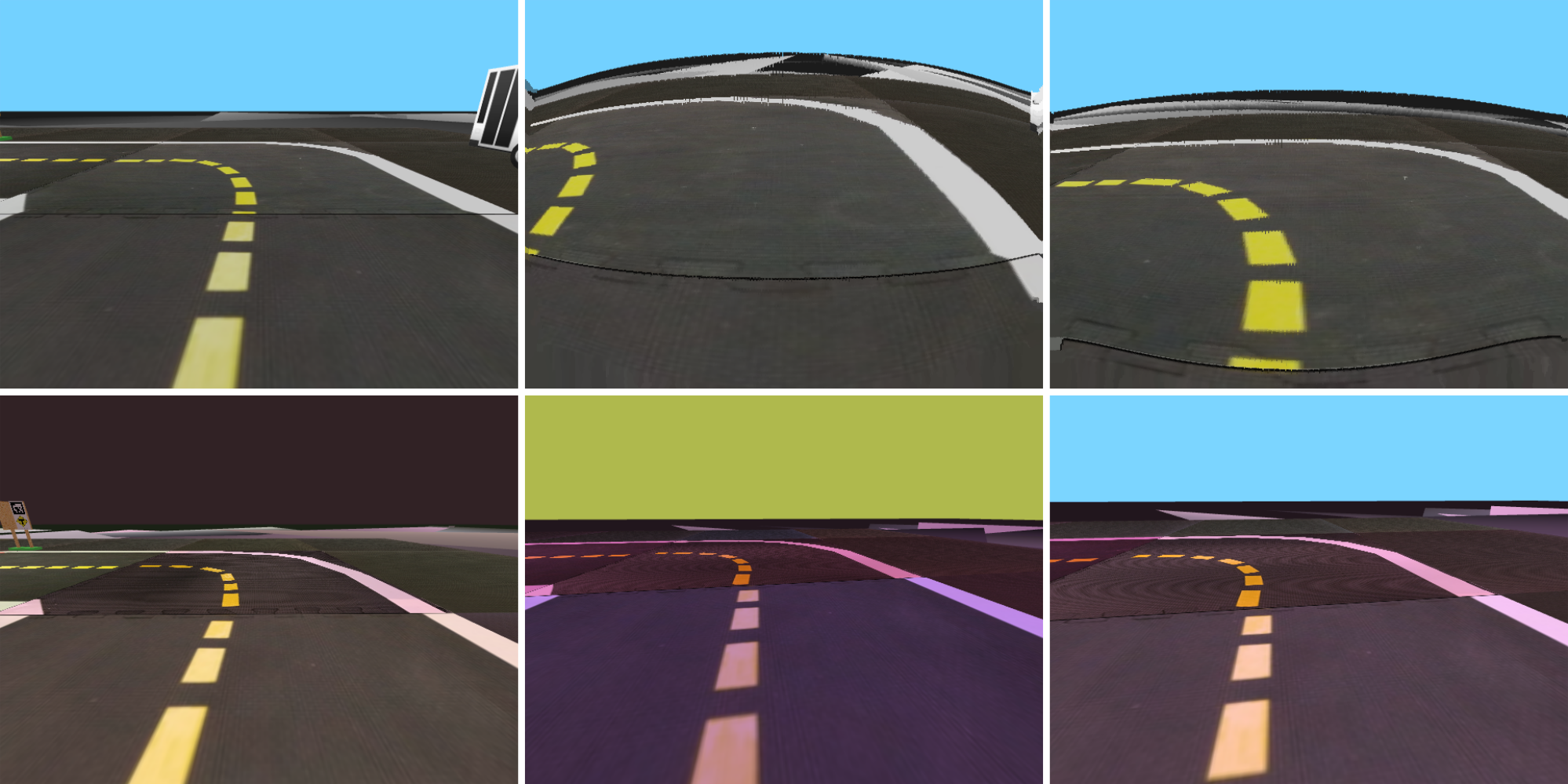}
	\caption{Examples of domain randomized observations}
	\label{fig_domain_rand}
\end{figure}

To train agents, we used an open-source simulation of the Duckietown environment\cite{GymDuckietown}. It models certain physical properties of the real environment accurately (dimensions of the robot, camera parameters, dynamic properties etc.), but several other effects (textures, objects surrounding the roads) and light simulation are less realistic (e.g. compared to modern computer games). These inaccuracies create a gap between simulation and reality which makes it challenging for any reinforcement learning agent to be trained in a simulation but operate in reality.  

To bridge the simulation to reality gap, and to achieve the generalization capability required for real performance we used domain randomization. This involves training the policy in many different variants of a simulated environment, by varying lighting conditions, object textures, camera, and vehicle dynamics parameters, road structures etc. (for examples of domain randomized observations see fig.~\ref{fig_domain_rand}). In addition to the "built-in" randomization options of Gym-Duckietown, we trained on a diverse set of maps to further improve the agent's generalization capability.

\subsection{Collision avoidance}
Collision avoidance with other vehicles greatly increases the complexity of the lane-following task. These problems can be solved in different ways, e.g. by overtaking or following from a safe distance. However, the sensing capability of the vehicle and the complexity of the policy determine the solution it can learn. Images from the forward-facing camera of a duckiebot only have $~160^\circ$ horizontal field of view, therefore the policy controlling the vehicle has no information about objects moving next to or behind the robot. Also, for simplicity, we chose a convolutional network and didn't incorporate an LSTM cell into it. For these reasons, it is unable to plan long maneuvers, such as overtaking, which also requires side-vision to check if returning to the right lane is safe. Therefore, we trained a policy in situations where there is a slow vehicle ahead, and the agent has to learn to perform lane following at full speed until it catches up with the vehicle upfront, then it must reduce its speed and keep a safe distance to avoid collision.

In these experiments, the \textit{Wheel Velocity - Braking} action mapping was used because this allows the policy to slow down or even stop the vehicle if necessary (unlike the one we call \textit{Steering}). Rewards used to train for collision avoidance were the modified version of the Orientation reward and Distance traveled (unchanged). The simulation we used provides a $p_{coll} $ penalty if the so-called safety circles of two vehicles overlap. The reward term calculated based on this penalty is proportional to its change if it's decreasing, otherwise, it's 0. 
\stepcounter{equation}
\begin{equation}
	r_{coll}=\left\{\begin{array}{ll}
	-\lambda_{coll}\cdot\Delta p_{coll} & \textrm{if } \Delta p_{coll}<0 \\ 
	0 & \textrm{otherwise} 
	\end{array}\right.
\end{equation}
This term is added to the Orientation reward and intends to encourage the policy to increase the distance from the vehicle ahead if it got too close. Collisions are only penalized by terminating the episode, without giving any negative reward.

\subsection{Evaluation} \label{sec_evaluation}

\begin{figure}[tbp]
	\centering
	\subfloat[Simulated]{
		\includegraphics[height=35mm]{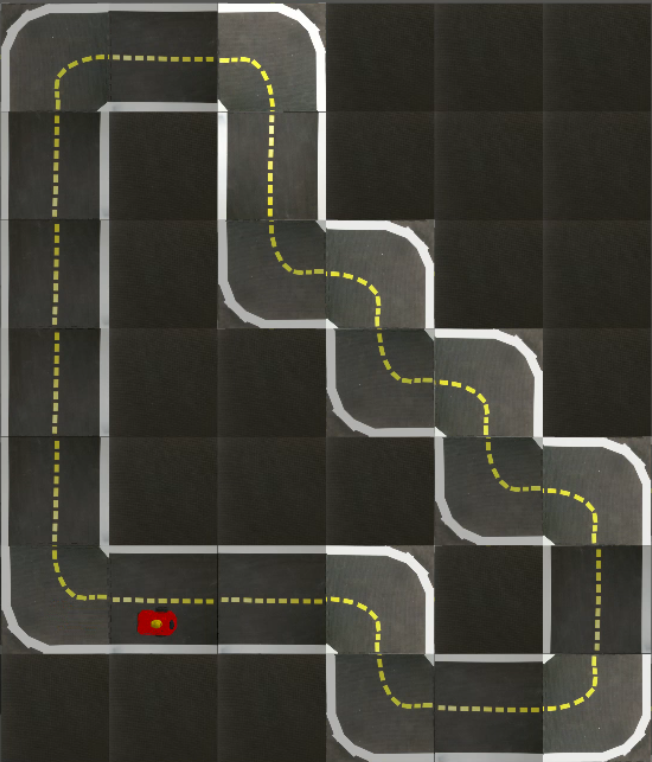}	\label{fig_test_track_sim}}
	\subfloat[Simulated]{
		\includegraphics[width=35mm,angle=90]{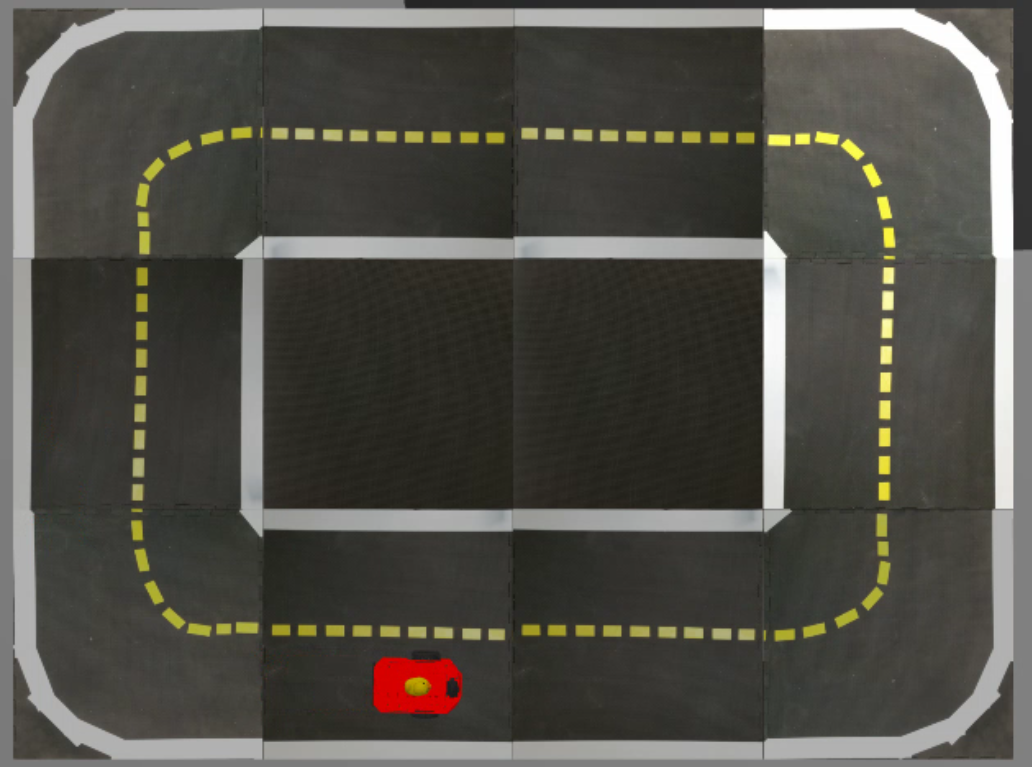}\label{fig_test_track_sim2real_sim}}
	\subfloat[Real]{
		\includegraphics[width=35mm,angle=90]{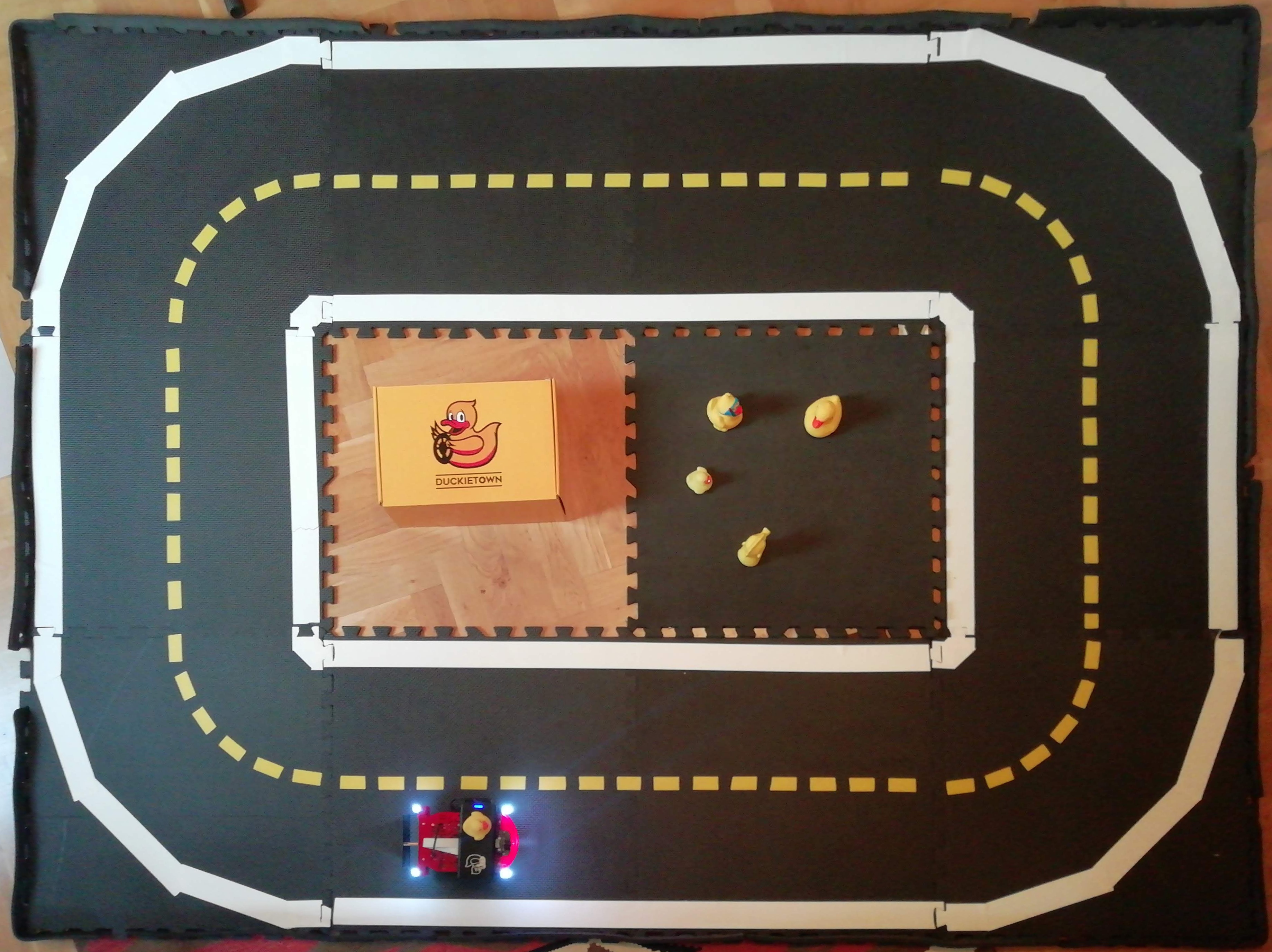}\label{fig_test_track_sim2real_real}}
	\caption{(a): Test track used for simulated reinforcement learning and baseline evaluations. (b),(c): Real and simulated test track used for the evaluation of the simulation to reality transfer.}
	\label{fig_test_track}
\end{figure}

To assess the performance of the reinforcement learning-based controller, we measured multiple performance metrics in the simulation and compared these against two baselines, one using a classical control theory approach, and human driving. To our knowledge no other methods have been published so far, which could be used as a baseline. 
These metrics are: 
\subsubsection{Survival time} The time until the robot left the road or the time period of an evaluation.
\subsubsection{Distance traveled in ego-lane [m]} The distance traveled along the right-hand-side lane under a fixed time period. Only longitudinal motion is counted, therefore tangential movement counts the most towards this metric.
\subsubsection{Distance traveled both lanes [m]} The distance traveled along the road under a fixed time period, but also sections where the agent moved in the oncoming lane count towards this metric. 
\subsubsection{Lateral deviation [m$\cdot$s]} Lateral deviation from the lane center line integrated over the time of an episode. 
\subsubsection{Orientation deviation [rad$\cdot$s]} The robot orientation's deviation from the tangent of the lane center line, integrated over the time of an episode. 

The classical control theory baseline relies on information about the robot’s relative location and orientation to the centerline of the lane, which are available in the simulator. %Gym-Duckietown implements the simulation of robots that preform lane following based on this information, our solution partially uses the code of these. 
This baseline works by controlling the robot to orient itself towards a point on it’s desired path ahead and calculating wheel velocities using a proportional-derivative (PD) controller, based on the orientation error of the robot. The parameters of this controller were hand-tuned to achieve sufficiently good performance, but more advanced control schemes could offer better results. 

In many reinforcement learning problems (e.g. the Atari 2600 games \cite{Atari2600}) the agents are compared to human baselines. Motivated by this benchmark we propose a method to measure how well humans are able to control duckiebots, which could be used as a baseline. The values shown in Table~\ref{tab_results_sim} were recorded by controlling the simulated robot using the arrow keys on a keyboard (therefore via discrete actions), while the observations seen by the human driver were very similar compared to the observations of the reinforcement learning agent.

\section{Results}

\begin{table}[tbp]
	\caption{Comparison of the reinforcement learning agent to two baselines in simulation}
	\begin{center}
		\begin{tabular}{lcccc}
			\hline
			Mean metrics over 5 episodes 						 & & RL           & PD        & Human  \\
			& & agent        & baseline  & baseline \\
			\hline
			Survival time [s]                  & $\uparrow$   & 15       & 15          & 15             \\
			Distance traveled both lanes [m]    & $\uparrow$   & 7.1     & 7.6          & 7.0           \\
			Distance traveled ego-lane [m]      & $\uparrow$   & 7.0     & 7.6          & 6.7           \\
			Lateral deviation [m$\cdot$s]      & $\downarrow$ & 0.5     & 0.5          & 0.9           \\
			Orientation deviation [rad$\cdot$s]& $\downarrow$ & 1.5     & 1.1          & 2.8           \\
			\hline
			%\multicolumn{5}{l}{For simulated evaluations the displayed values are means from 5 episodes, recorded on the same test map, from 5 different starting positions.}
			
		\end{tabular}
		\label{tab_results_sim}
	\end{center}
\end{table}

\begin{figure}[tbp]
	\centering
	\includegraphics[width=6cm]{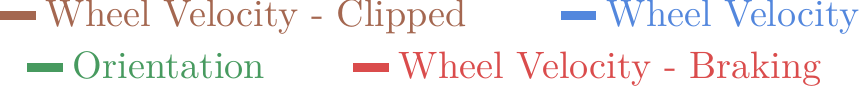}\\ \vspace{-3mm}
	\subfloat[Orientation reward]{\includegraphics[height=3.1cm]{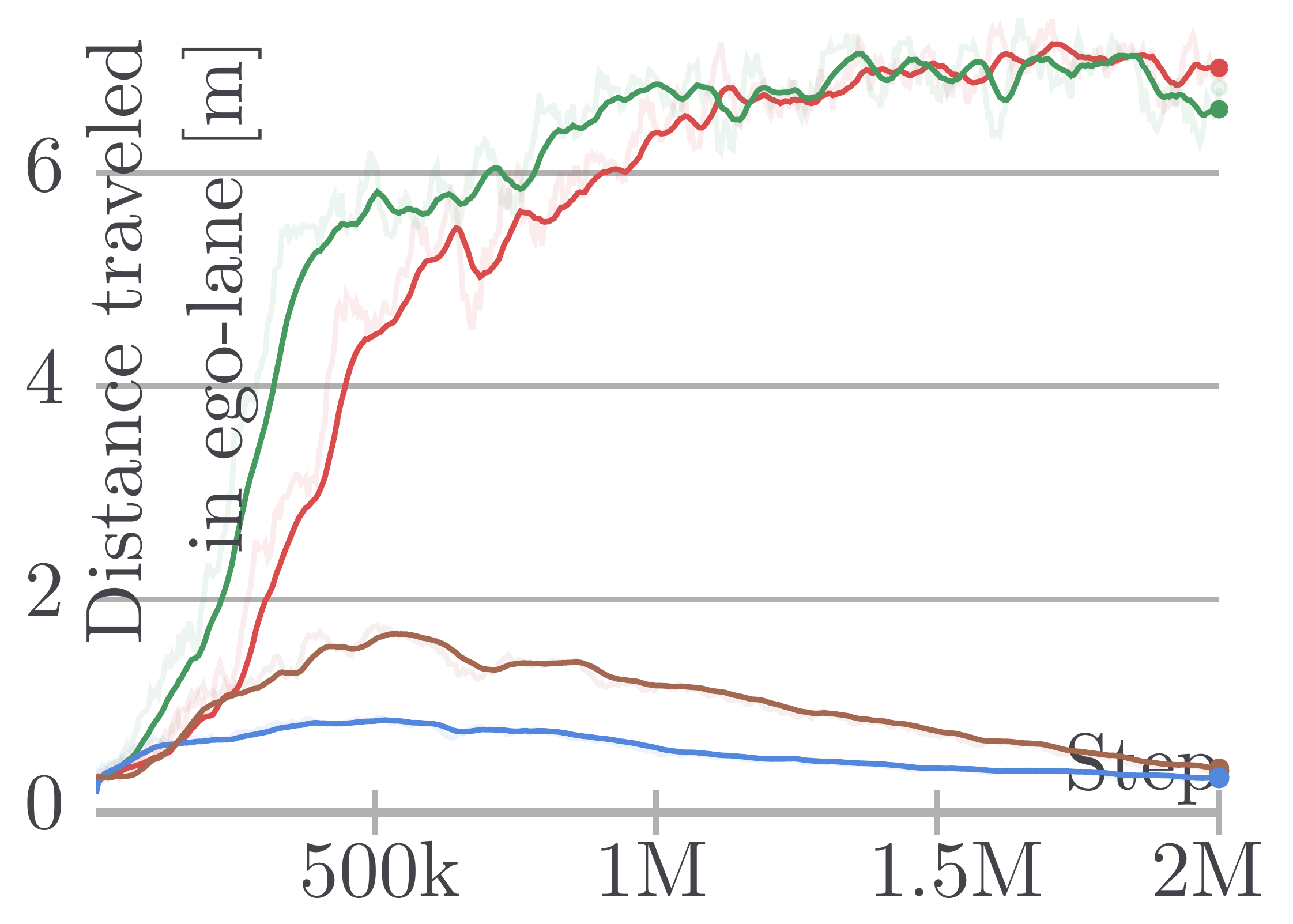}}	
	\subfloat[Distance travelled reward]{\includegraphics[height=3.1cm]{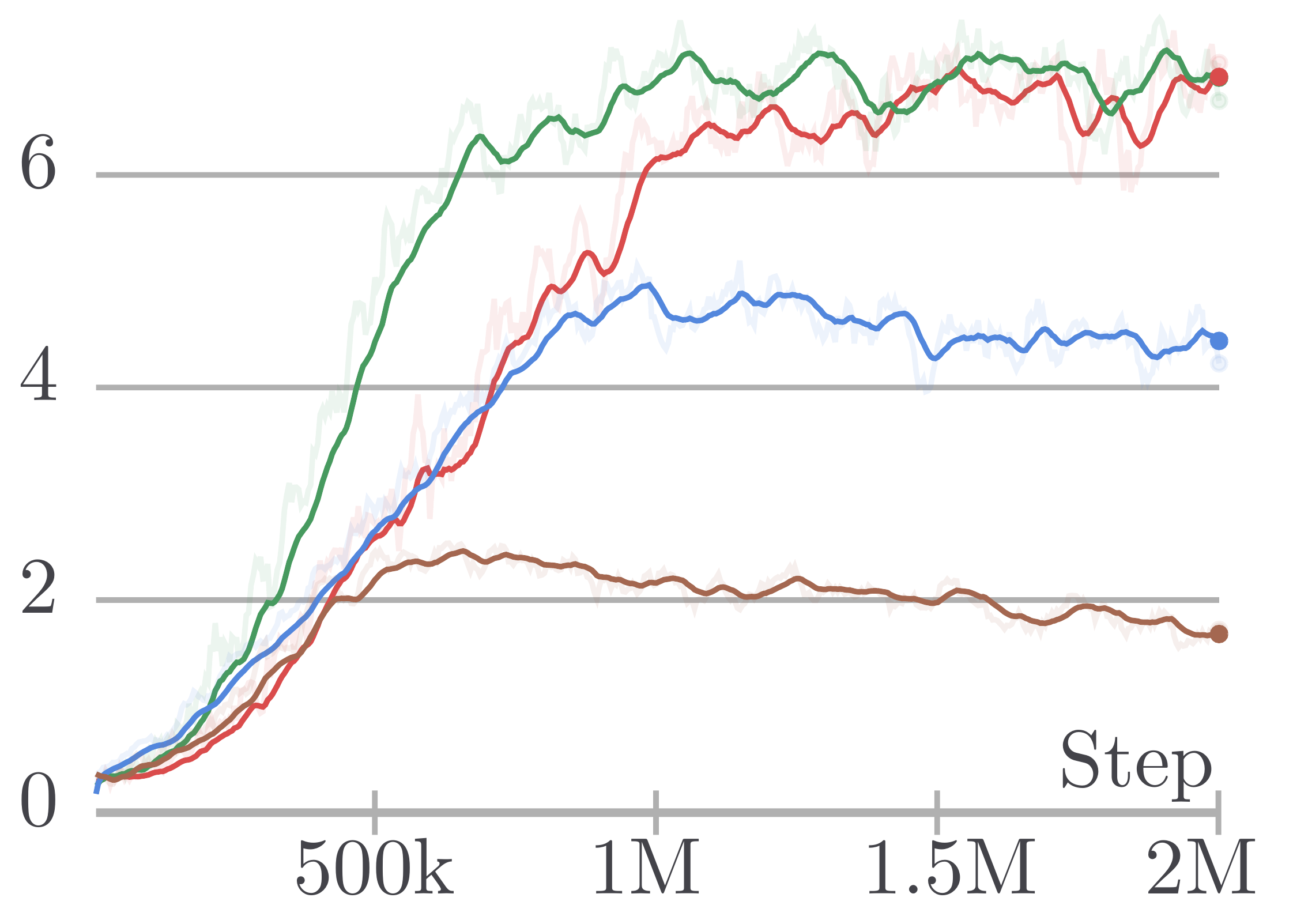}}
	\caption{Learning curves for the reinforcement learning agent with different action representations and reward functions.}
	\label{fig_rewards_actionmappings}
\end{figure}

\begin{figure*}[!ht]
	\centering
	\subfloat[{$t=0$[s] Initial positions}]{\includegraphics[width=27mm]{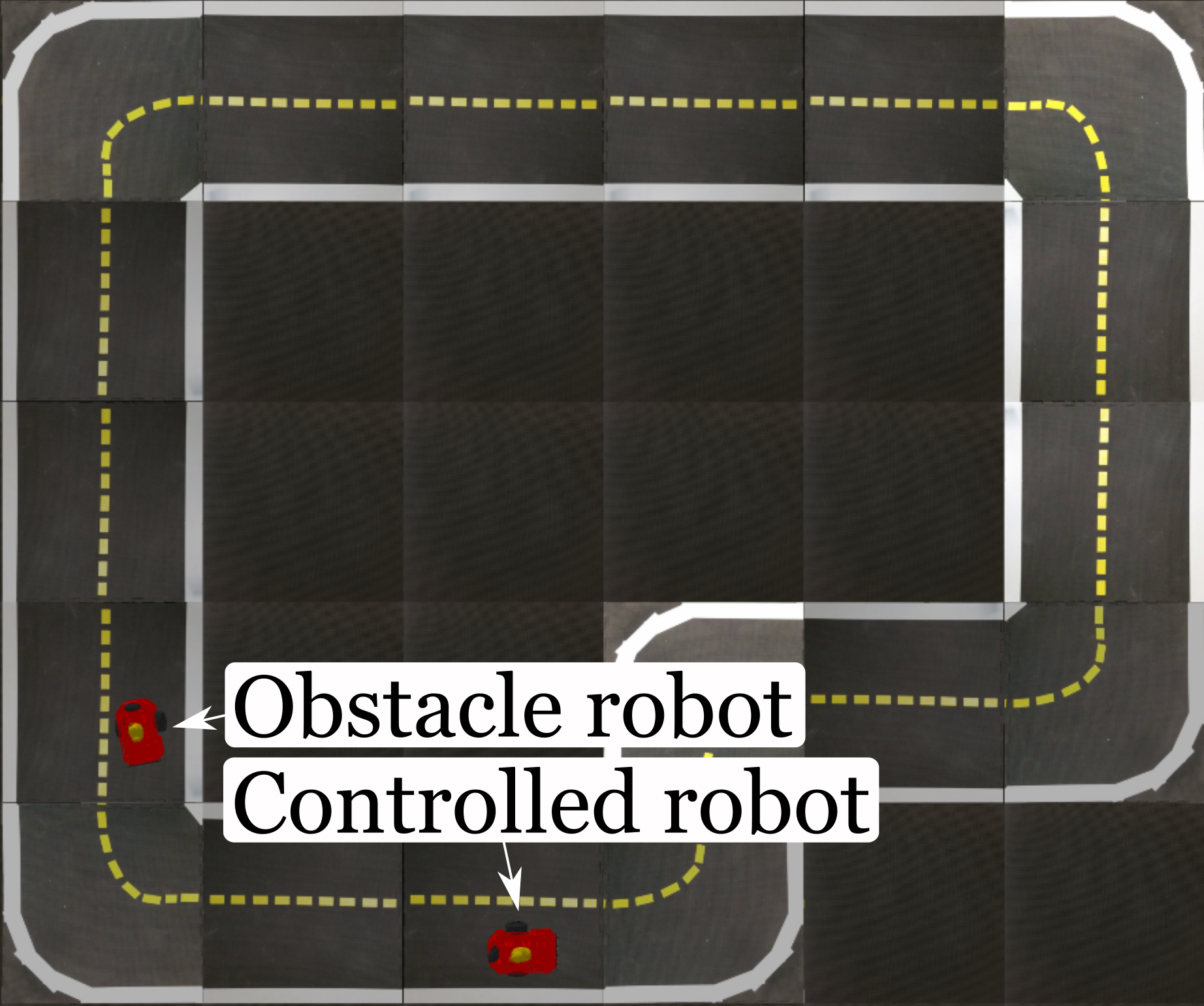}} \enspace
	\subfloat[{$t=6$[s] Catching up}]{\includegraphics[width=27mm]{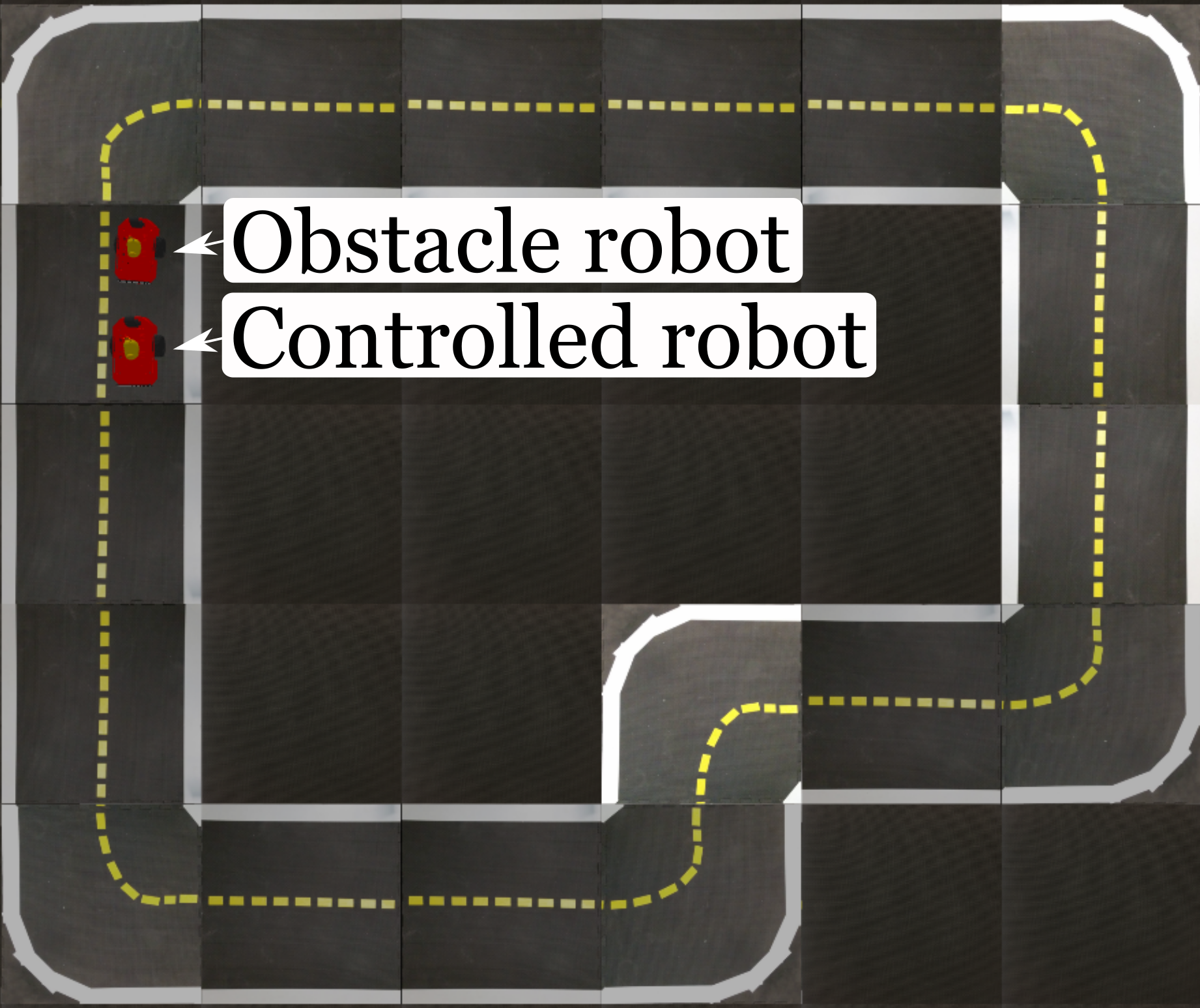}} \enspace
	\subfloat[{$t=8$[s]}]{\includegraphics[trim=842pt 595pt 842pt  595pt, clip,width=27mm]{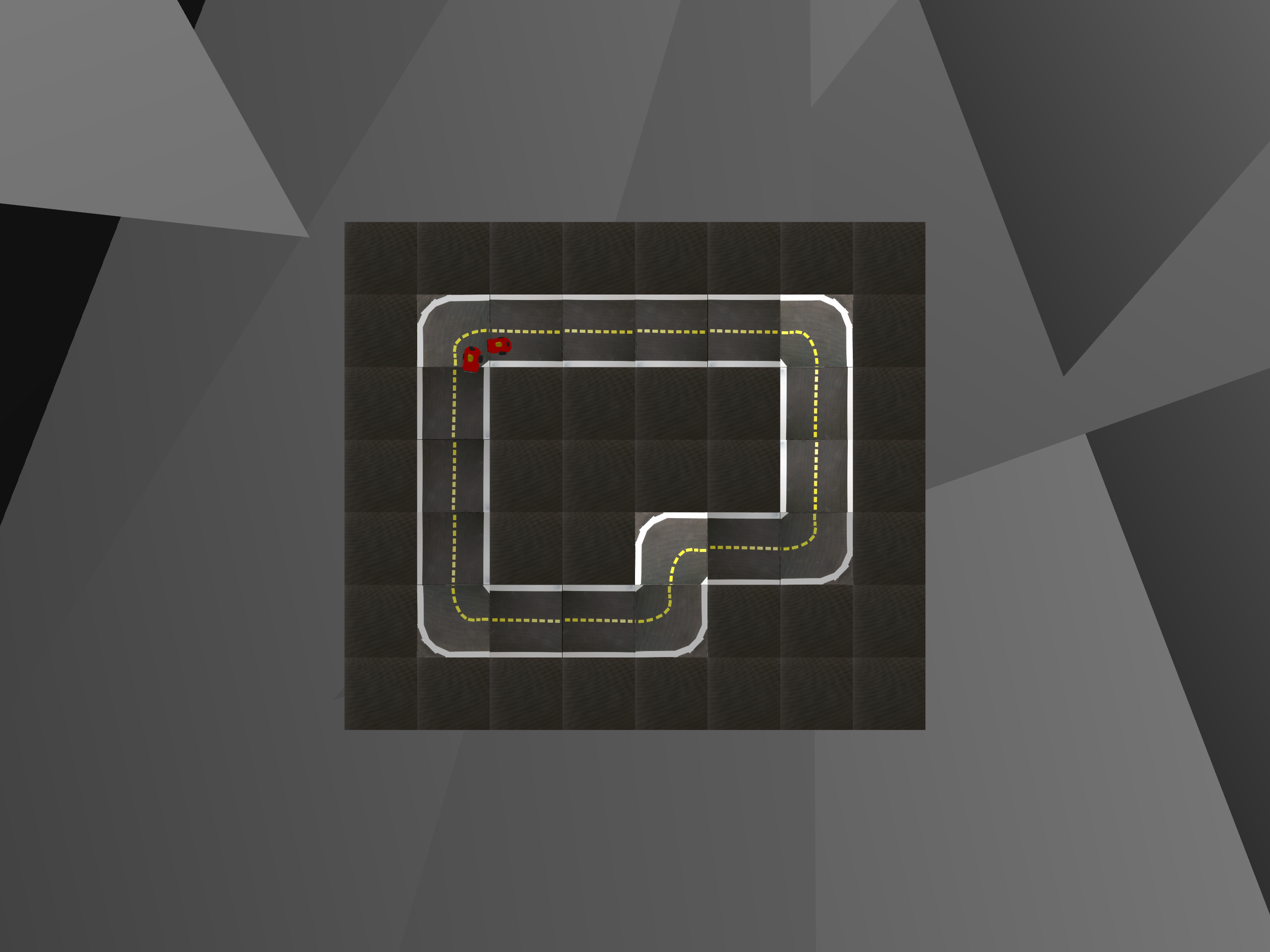}} \enspace
	%	\subfloat[{$t=15$[s]}]{\includegraphics[trim=842pt 595pt 842pt  595pt, clip,width=27mm]{TrajectoryLFV_TopView_t15s}} \enspace
	%	\subfloat[{$t=20$[s]}]{\includegraphics[trim=842pt 595pt 842pt  595pt, clip,width=27mm]{TrajectoryLFV_TopView_t20s}} \enspace
	\subfloat[{$t=24$[s]}]{\includegraphics[trim=842pt 595pt 842pt  595pt, clip,width=27mm]{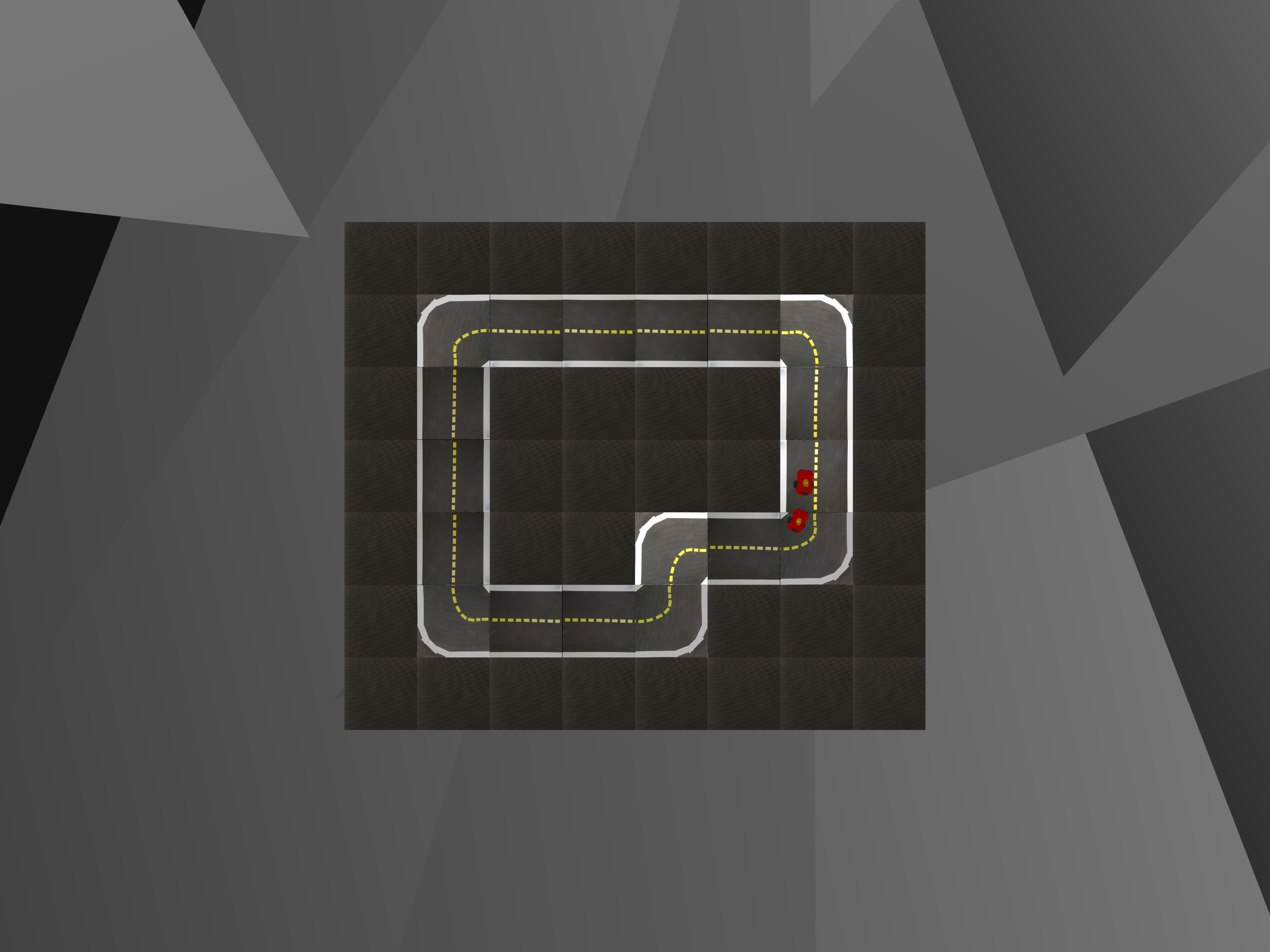}} \enspace
	\subfloat[Approximate distance between the vehicles.]{\includegraphics[width=58mm]{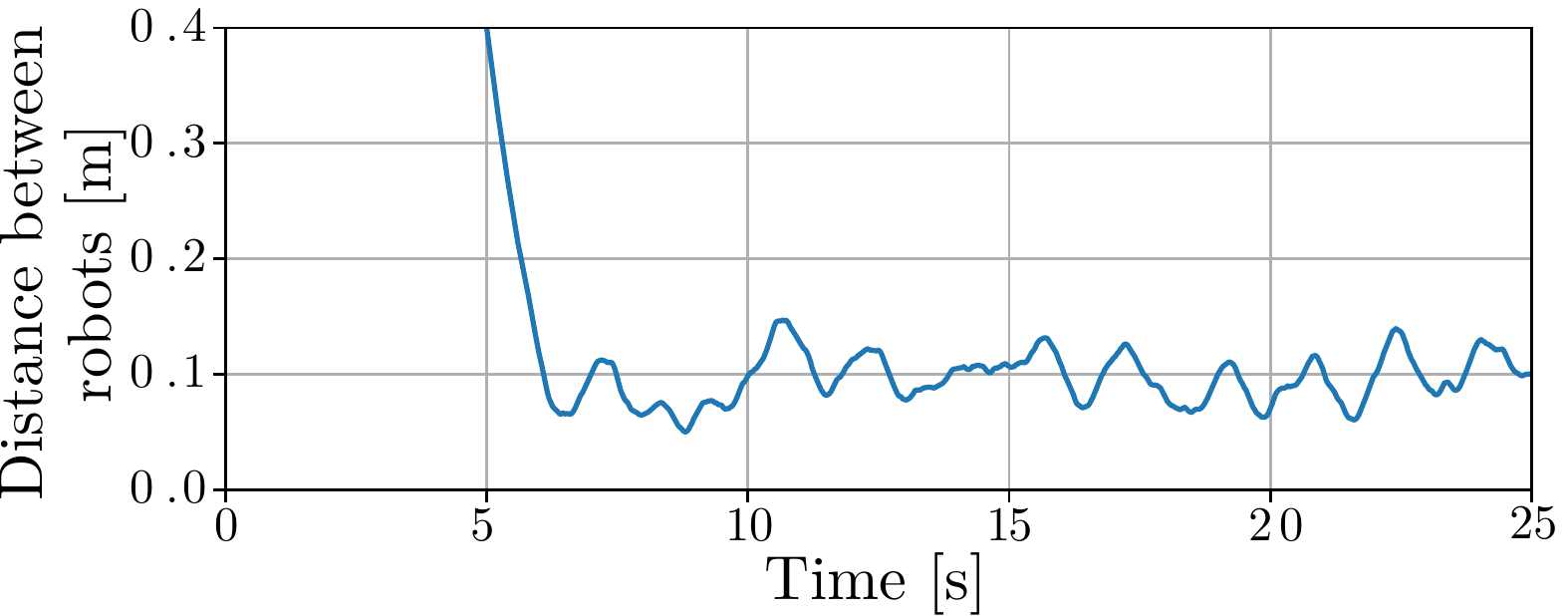}\label{fig_lfv_distance_plot}}
	\caption{Sequence of robot positions in a collision avoidance experiment with a policy trained using the modified Orientation reward. After $t=6[s]$ the controlled robot follows the vehicle in front of it from a short, but safe distance until the end of the episode. (Approximate distance is calculated as the distance between the center points of the robots minus the length of a robot.)}
	\label{fig_lfv_traj_sequence}
\end{figure*}

\subsection{Simulation}
Even though multiple papers demonstrate the feasibility of training vision-based driving policies using reinforcement learning, adapting to a new environment still poses many challenges. Due to the high dimensionality of the image-like observations, many algorithms converge slowly and are very sensitive to hyperparameter selection. Our method, using Proximal Policy Optimization is able to converge to good lane following policies in 1 million timesteps, thanks to the high sample-complexity of the algorithm.

\subsubsection{Comparing against baselines}Table~\ref{tab_results_sim} compares our reinforcement learning agent to the baselines. The performance of the trained policy is measurable to our classical control theory baseline, as well as to how well humans are able to control the robot in the simulation. Most metrics indicate similarly good or equal performance, even though the PD controller baseline relies on high-level data such as position and orientation error, rather than images. 

\subsubsection{Action representation and reward shaping} Experiments with different action representations show that constrained and preferably biased action spaces allow convergence to good policies (\textit{Wheel Velocity - Braking} and \textit{Steering}), however, more general action spaces (\textit{Wheel Velocity} and it's \textit{Clipped} version) can only converge to inferior policies under the same number of steps (see fig.~\ref{fig_rewards_actionmappings}). The proposed orientation based reward function also leads to as good final performance as "trivially" rewarding based on the distance traveled, however, the latter seems to perform better on more general action representations (because policies using these action spaces and trained with the Orientation reward doesn't learn to move fast).

\subsection{Real-world driving}

\begin{table}[tbp]
	\caption{Evaluation results of reinforcement learning agent in the real environment and in matching simulations}
	\begin{center}
		\begin{tabular}{llccc}
			\hline
			Eval.  & Mean metrics over 6 episodes    &            &          Domain           & Nominal \\
			Domain &                                 &            &           Rand.           &         \\
			\hline
			Real   & Survival time [s]               & $\uparrow$ &           54              &  45     \\
			       & Distance traveled both lanes [m] & $\uparrow$ &           15.6            &  11.4   \\
			       & Distance traveled ego-lane [m]   & $\uparrow$ &           7.0             &  8.4    \\ 
		    \hline
			Sim.   & Survival time [s]               & $\uparrow$ &            60             &   60    \\
			       & Distance traveled [m]            & $\uparrow$ &           15.5            &  15.0   \\ 
			\hline& 
		\end{tabular}
		\label{tab_sim2real}
	\end{center}
\end{table}
To measure the quality of the transfer learning process and the performance of the controller in the real world, we selected performance metrics that are easily measurable both in reality and simulation. These were recorded in both domains in matching experiments and compared against each other. The geometry of the tracks, the dimensions, and speed of the robot are simulated accurately enough, to evaluate the robustness of the policy against all inaccurately and not simulated effects. Using this method, we tested policies trained in the domain randomized simulation, but also ones that were trained only in the "nominal" simulation. This allows us to evaluate the transfer learning process and highlight the effects of training with domain randomization. The real and simulated version of the test track used in this analysis is shown on fig.~\ref{fig_test_track_sim2real_sim}~and~\ref{fig_test_track_sim2real_real}. 

During real evaluations, generally, we experienced that under ideal circumstances (no distracting objects outside the roads and good lighting conditions) the policy trained in the "nominal" simulation is able to drive reasonably well. However, training with domain randomization leads to more reliable robust performance in the real world.

Table~\ref{tab_sim2real} show the quantitative results of this evaluation. The two policies seem to perform equally well if comparing them based on their performance in the simulation. However, metrics recorded in the real environment show that the policy trained with domain randomization performs almost as well as in the simulation, while the other policy performs noticeably worse. The lower \textit{Distance traveled ego-lane} metric of the domain randomized policy is because the vehicle tends to drift to the left lane in sharp turns but returns to the right-lane afterward, while the nominal policy usually made more serious mistakes. Note that in these experiments the Orientation based reward and the Steering action representation were used, as this configuration learns to control the robot in the least amount of steps and training time.

An online video demonstrates the performance of our trained agent: \url{https://youtu.be/kz7YWEmg1Is}

\subsection{Collision avoidance}
\begin{table}[tbp]
	\caption{Evaluation results of policies trained for collision avoidance with different reward functions}
	\begin{center}
		\begin{tabular}{lccc}
			\hline
			Mean metrics over 15 episodes 	   &              & Distance & Orientation \\
			                                   &              & traveled &  $+ r_{coll}$    \\
			\hline
			Survival time (max. 60) [s]        & $\uparrow$   & 46       & 52   \\
			Distance traveled both lanes [m]    & $\uparrow$   & 22.5     & 22.9 \\
			Distance traveled ego-lane [m]      & $\uparrow$   & 22.7     & 23.1 \\
			Lateral deviation [m$\cdot$s]      & $\downarrow$ & 1.9      & 1.6  \\
			Orientation deviation [rad$\cdot$s]& $\downarrow$ & 6.3      & 5.8  \\
			\hline
		\end{tabular}
		\label{tab_results_collsion}
	\end{center}
	\vspace*{-3mm}
\end{table}

Fig~\ref{fig_lfv_traj_sequence} demonstrates the learned collision avoidance behavior. In the first few seconds of the simulation, the robot controlled by the reinforcement learning policy accelerates to full speed. Then, as it approaches the slower, non-learning robot, it reduces it's speed and maintains approximately a constant distance from the vehicle ahead (see fig~\ref{fig_lfv_distance_plot}). 

Table~\ref{tab_results_collsion} shows that training with both reward functions lead to functional lane-following behavior, however the non-maximal \textit{Survival time} values indicate that neither of the policies are capable of performing lane following reliably with the presence of an obstacle robot for 60 seconds. All metrics in Table~\ref{tab_results_collsion} indicate that the modified Orientation reward leads better lane following metrics, than the simpler Distance traveled reward. It should be noted, that these metrics were mainly selected to evaluate the lane following capabilities of an agent, more in-depth analysis of collision avoidance with a vehicle upfront call for more specific metrics.
  
An online video demonstrates the performance of our trained agent: \url{https://youtu.be/8GqAUvTY1po}

\enlargethispage{-2.8cm}

\subsection{Salient object maps}
Visualizing which parts of the input image contribute the most to a particular output (action) is important, because, it provides some explanation of the network's inner workings. Fig.~\ref{fig_sal_obj} shows salient object maps in different scenarios, generated using the method proposed in \cite{SalientObj}. All of these images indicate high activations on lane markings, which is expected.

\section{Conclusions}
This work presented a solution to the problem of complex, vision-based lane following in the Duckietown environment using reinforcement learning to train an end-to-end steering policy capable of simulation to real transfer learning. We found that the training is sensitive to problem formulation, for example to the representation of actions.
We showed that by using domain randomization, a moderately detailed and accurate simulation is sufficient for training end-to-end lane following agents that operate in a real environment. The performance of these agents was evaluated by comparing some basic metrics in matching real and simulated scenarios.
Agents were also successfully trained to perform collision avoidance in addition to lane following. 
Finally, salient object visualization was used to give an illustrative explanation of the inner workings of the policies, in both the real and simulated domains.

%Solving vision-based end-to-end driving using reinforcement learning is a challenging task and 

\section*{Acknowledgment}
We would like to show our gratitude to professor Bálint Gyires-Tóth (BME, Dept. of Telecommunications and Media Informatics) for his assistance and comments on the progress of our research. 

The research reported in this paper and carried out at the Budapest University of Technology and Economics was supported by Continental Automotive Hungary Ltd. and the “TKP2020, Institutional Excellence Program” of the National Research Development and Innovation Office in the field of Artificial Intelligence (BME IE-MI-SC TKP2020).

\bibliography{ismcr}{}
\bibliographystyle{ieeetr}

\end{document}